\documentclass[10pt,twocolumn,letterpaper]{article}

\usepackage{cvpr}              % To produce the CAMERA-READY version
\usepackage{graphicx}
\usepackage{amsmath}
\usepackage{amssymb}
\usepackage{booktabs}
\usepackage{bm}
\usepackage{dsfont}
\usepackage{enumitem}
\usepackage{graphicx}
\usepackage{algorithm}
\usepackage{algpseudocode}
\usepackage{ctable}
\usepackage{comment}
\usepackage[misc]{ifsym}
\usepackage[pagebackref,breaklinks,colorlinks]{hyperref}

\usepackage[capitalize]{cleveref}
\crefname{section}{Sec.}{Secs.}
\Crefname{section}{Section}{Sections}
\Crefname{table}{Table}{Tables}
\crefname{table}{Tab.}{Tabs.}

\def\cvprPaperID{3114} % *** Enter the CVPR Paper ID here
\def\confName{CVPR}
\def\confYear{2022}
\newtheorem{prop}{Proposition}

\begin{document}

\title{DTFD-MIL: Double-Tier Feature Distillation Multiple Instance Learning for Histopathology Whole Slide Image Classification}

\author{Hongrun Zhang$^{1}$, Yanda Meng$^{1}$, Yitian Zhao$^{2}$, Yihong Qiao$^{3}$, Xiaoyun Yang$^{4}$, Sarah E. Coupland$^{1}$ \\ Yalin Zheng$^{1}$\Letter\\ 
$^{1}$University of Liverpool, $^{2}$Cixi Institute of Biomedical Engineering, Chinese Academy of Sciences\\
$^{3}$China Science IntelliCloud Technology Co., Ltd, $^{4}$Remark AI UK Limited, London \\
{\tt\small  \{hongrun.zhang,   yanda.meng, S.E.Coupland, yalin.zheng\}@liverpool.ac.uk } \\
{\tt\small yitian.zhao@nimte.ac.cn,  yihong.qiao@intellicloud.ai,  xyang@remarkholdings.com} 
%\and
}
\maketitle

%%%%%%%%% ABSTRACT
\begin{abstract}
Multiple instance learning (MIL) has been increasingly used in the classification of histopathology whole slide images (WSIs). However, MIL approaches for this specific classification problem still face unique challenges, particularly those related to small sample cohorts. In these, there are limited number of WSI slides (bags), while the resolution of a single WSI is huge, which leads to a large number of patches (instances) cropped from this slide. To address this issue, we propose to virtually enlarge the number of bags by introducing the concept of pseudo-bags, on which a double-tier MIL framework is built to effectively use the intrinsic features. Besides, we also contribute to deriving the instance probability under the framework of attention-based MIL, and utilize the derivation to help construct and analyze the proposed framework. The proposed method outperforms other latest methods on the CAMELYON-16 by substantially large margins, and is also better in performance on the TCGA lung cancer dataset. The proposed framework is ready to be extended for wider MIL applications. The code is available at: \url{https://github.com/hrzhang1123/DTFD-MIL}

%\color{blue}{The code will be released.}\footnote{will be available at Github upon acceptance.} 

%Multiple instance learning (MIL) is a difficult task, yet it is of great utility. MIL on histopathology whole slide images is even more challenging, as there are limited number of slides (bags) while a great number of patches (instances) in each slide. To alleviate this issue, we propose to virtually enlarge the number of bags by introducing the concept of pseudo-bags; and building up a double-tier framework to compensate for the drawback of the pseudo-bags. Besides, we also contribute to deriving the instance probability under the framework of attention-based MIL which is the basic formulation of the latest and most advanced MIL methods, and utilize the derivation to help construct as well as analyze the proposed framework. Shown from the experimental results, the proposed method outperforms other latest works on the CAMELYON-16 with substantially large margins, including a recent concurrent work, and is significantly better in performance on the TCGA lung cancer dataset. \color{blue}{The code will be released.} 
   
\end{abstract}

%%%%%%%%% BODY TEXT
\section{Introduction}
\label{sec:intro}

The automation of whole slide images (WSIs) poses a significant challenge to the field of computer vision. The increasing use of WSIs in histopathology results in digital pathology providing huge improvements in workflow and diagnosis decision-making by pathologists \cite{pantanowitz2011review, cornish2012whole, litjens2016deep, madabhushi2009digital, pinckaers2020streaming}, but it also stimulates the need  for intelligent or automatic analytical tools of WSIs \cite{he2012histology, wang2016deep, tellez2019neural, zhang2020piloting, li2018cancer, zhang2021regularization, wang2021predicting}. WSIs have enormous sizes, ranging from 100M pixels to 10G pixels, and this unique characteristic makes it almost infeasible to directly transfer existing machine learning techniques to their applications, since these existing techniques were initially intended for natural images or  medical images with much smaller sizes. When it comes to deep learning based models, large scale datasets and high quality annotations are the primary yet crucial conditions to train a high capacity model. However, the enormous sizes of WSIs bring along substantial burden for pixel-level annotation. This problem in turn encourages researchers to develop deep learning based models trained with limited annotations, termed as ``Weakly Supervised" or ``Semi-Supervised" \cite{wang2019weakly, lu2019semi, Meng_2021_ICCV, sharma2021cluster}. A large proportion of existing weakly supervised works for WSI classification are characterized as ``multiple instance learning" (MIL) \cite{amores2013multiple, dietterich1997solving, maron1998framework, chen2013multi}. Under the framework of MIL, a slide (or WSI), acting as a bag, constitutes multiple instances that are hundreds or thousands of patches cropped from the slide. With at least one instance being disease positive, the slide is marked as  positive, or otherwise  negative.

\begin{figure}[t]
	\centering
	\includegraphics[width=0.4\textwidth]{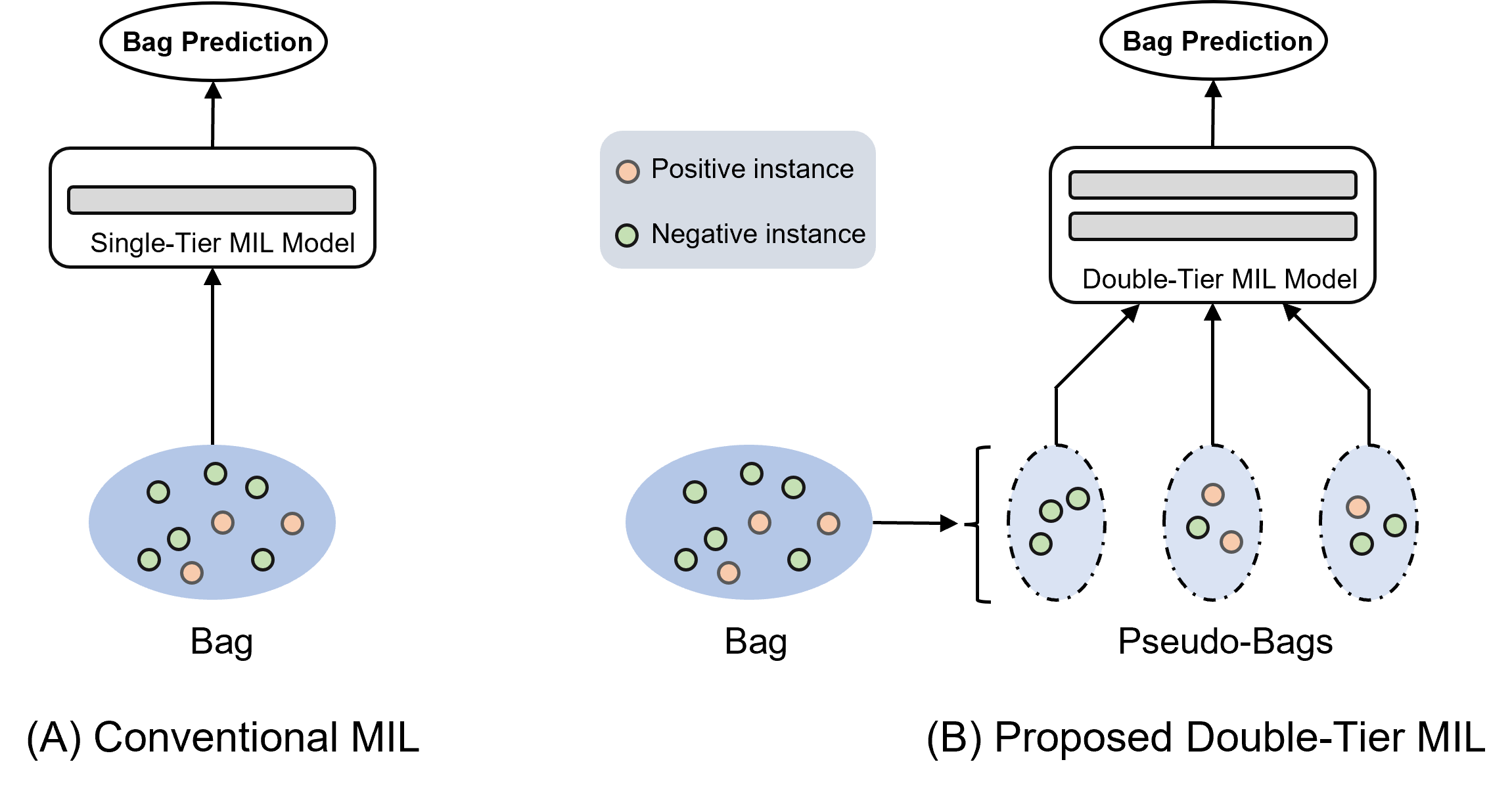}
	\caption{ Illustration of the difference between conventional MIL models and the proposed double-tier MIL model. }
	\label{fig_compare}
\end{figure}

\begin{figure*}[ht]
	\centering
	\includegraphics[width=0.7\textwidth]{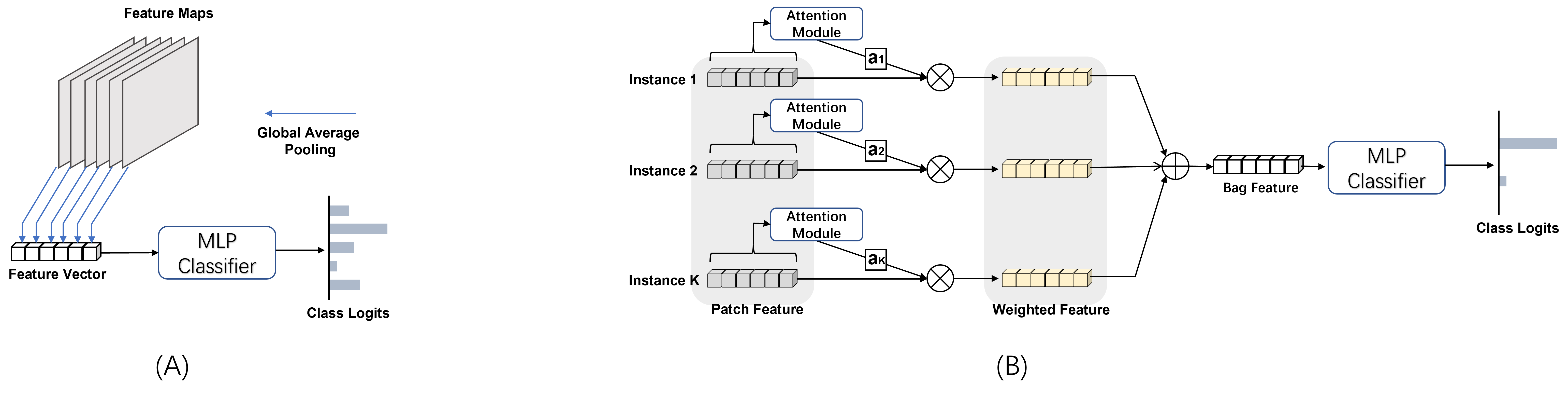}
	\caption{(A) Illustration of an image classification system of deep learning. Global average pooling is applied to the extracted feature maps of an image, leading to a feature vector representing the image. Then the feature vector is forwarded to a classifier which outputs the class logits, then class probabilities by softmax. (B) Illustration of the AB-MIL paradigm. The extracted feature of an instance is weighted by an attention score. The bag feature is obtained through the sum of the weighted instance features, and then fed into a classifier for the bag prediction. }
	\label{fig_class_gradcam}
\end{figure*}

There exist some successful attempts to solve the MIL problem in various computer vision tasks \cite{li2019multi, oquab2014weakly, pathak2014fully, pinheiro2015image, oquab2015object}. The innate characteristics of WSIs, however, make it less straightforward to develop MIL solution for WSI classification than the counterparts in other computer vision sub-fields, as the only direct guidance information for training are the labels of a few hundred slides. The most notorious consequence is the over-fitting problem where a machine learning model tends to fall into local minima during optimization while the learnt  features are less relevant with respect to  target disease, and as a result, the trained model has inferior generalization ability.

Most recent works of MIL for WSIs to tackle the overfitting issue  are built on the essential idea of exploiting more information to learn in addition to the  labels of the relatively small number of slides in a cohort. The mutual-instance relation is an important direction to explore for this purpose, and it has been empirically proven to be effective. The mutual-instance relation can be specified as spatial \cite{chikontwe2020multiple} or feature distances \cite{li2021dual, xie2020beyond, sharma2021cluster, tu2019multiple}, or can be agnostically learnt by neural modules, such as recurrent neural networks (RNN) \cite{campanella2019clinical}, transformer \cite{shao2021transmil}, and graph convolution network \cite{zhao2020predicting}.

Many of the aforementioned methods belong to attention-based MIL (AB-MIL) \cite{ilse2018attention}, although they differ in the formulations of the attention scores. However, it was believed infeasible to explicitly infer instance probabilities under AB-MIL frameworks  \cite{li2021dual}, and as an alternative, attention scores were usually used as the indications of positive activation \cite{ilse2018attention, li2021dual, shao2021transmil, godson2022weaklysupervised}. In this paper, we argue that the attention score is not a rigorous metric for this purpose, and instead  we contribute to derive the instance probability under the framework of AB-MIL.

Given the huge size of a WSI, the units to be directly processed are the much smaller patches cropped from  WSIs \cite{hou2016patch}.  MIL models for WSI classification essentially aim to recognize the most distinctive patches that correspond mostly to the slide label. However, there are a limited number of slides while there are hundreds or even thousands of patches (instances) in a slide, and the information for learning are only the slide-level labels. Moreover, in many histopathology slides the positive regions corresponding to positive diseases only occupy small portions of tissue, leading to a small ratio of positive instances of a slide. Therefore, it is challenging to guide a model to recognize positive instances under the condition of MIL, since these factors collectively contribute to deteriorating the over-fitting problem.

\begin{figure*}[ht]
	\centering
	\includegraphics[width=0.8\textwidth]{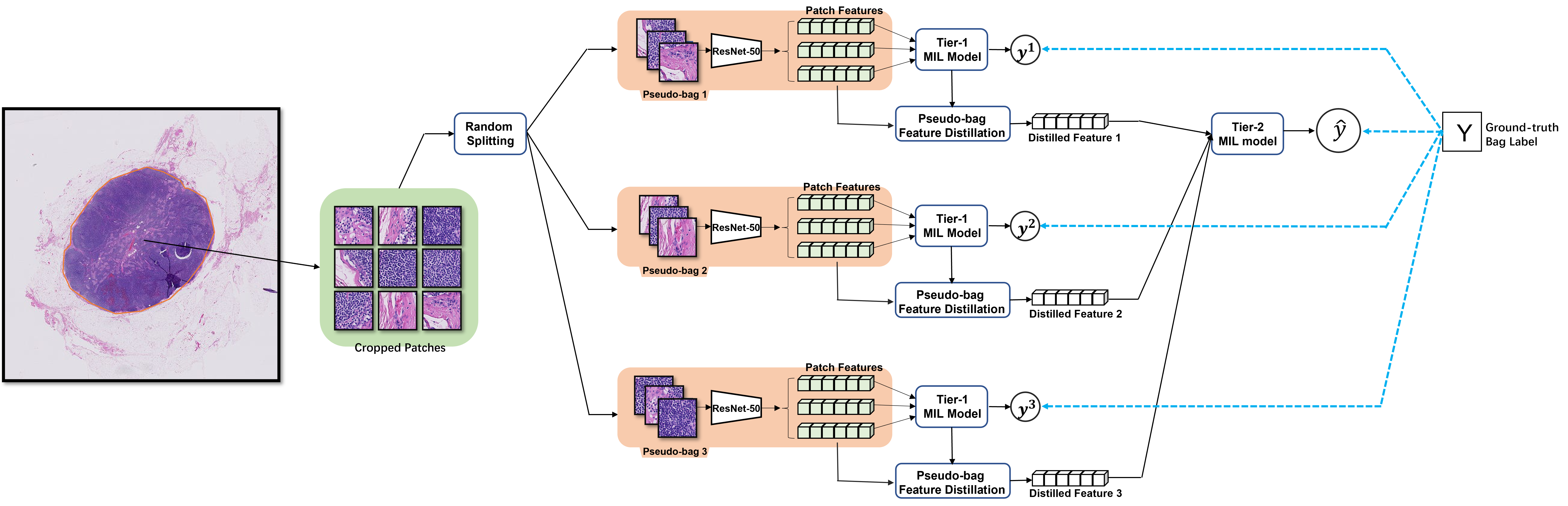}
	%\includesvg[width=0.92\textwidth]{image/overview3.svg}
	\caption{Overview of the proposed DTFD-MIL. A set of patches (we show only 9 for convenience) are first cropped from the tissue regions of a slide. these patches are randomly split into $M$ pseudo-bags ($M=3$ for example). A tier-1 MIL model is then applied to the 3 pseudo-bags, respectively. Based on the outputs of the Tier-1 model on the 3 pseudo-bags, 3 feature vectors are distilled accordingly and are then forwarded to the Tier-2 MIL model. The ground-truth bag label supervises both the Tier-1 and Tier-2 models during the training, denoted by blue dash lines.}
	\label{fig_overview}
\end{figure*}

Although most recent methods utilize mutual-instance relations to improve MIL, they do not explicitly confront the problems originated from the innate characteristics of WSIs as mentioned above. To alleviate the negative impacts of these problems, we introduce the concept of `pseudo-bag' in the proposed framework. That is, we randomly split the instances (patches) of a bag (slide) into several smaller bags (pseudo-bags), and assign each pseudo bag the label of the original bag, termed as the parent bag. This strategy virtually increases the number of bags while inside each pseudo bag there are fewer instances; it also enables Double-Tier Feature Distillation MIL model (Fig.\ref{fig_compare}). More specifically, a Tier-1 AB-MIL model is applied to the pseudo-bags of all the slides. However, it comes along the risk that a pseudo-bag from a positive parent bag may not be allocated with at least one positive instance, in which case a mislabeled pseudo-bag is introduced. To tackle this issue, we distill a feature vector from each pseudo-bag and establish a Tier-2 AB-MIL model upon the features distilled from all the pseudo-bags of a slide (See Fig.\ref{fig_overview}). Through the distillation process, the Tier-1 model provides initial candidates of distinct features for the Tier-2 model to generate a better representation for the corresponding parent bag. Furthermore, for the purpose of feature distillation, we derive the instance probability under the framework of AB-MIL, by harnessing  the fundamental idea of Grad-CAM that was  developed for visualizing deep learning features \cite{selvaraju2017grad}.

In essence, we deal with the MIL problem for WSI classification from another perspective with the proposed double-tier MIL framework. The main contributions are: (1).  We introduce the concept of pseudo-bags to alleviate the issue of limited number of WSIs. (2) We derive the instance probability under the AB-MIL framework by utilizing the essential idea of Grad-CAM. Given AB-MIL is the base for many MIL works, the instance probability derivation can help with the extension studies of related MIL methods. (3). By utilizing the instance probability derivation, we formulate a double-tier MIL framework, and the experiments show its superiority to other latest methods on two large public histopathology WSI datasets.

\section{Related Works}
\subsection{Multiple instance learning  in WSI analysis}

Providing the importance of weakly supervised learning, there is a trend to develop MIL algorithms for WSI analysis where, instead of elaborated pixel-level annotation, only the slide labels are available for training. MIL models generally can be  divided into two groups, based on whether the final bag predictions are directly from  instance predictions \cite{feng2017deep, campanella2019clinical, hou2016patch, kanavati2020weakly, xu2019camel, lerousseau2020weakly}, or from the aggregations of features of instance \cite{zhu2017deep, wang2018revisiting, lu2021data, ilse2018attention, shao2021transmil, li2021dual, sharma2021cluster}. For the former, the bag predictions are usually obtained through either the average pooling (mean value of probabilities of instances) or the maximum pooling (max value of probabilities of instances). In contrast, the latter one learns a high-level representation of a bag and builds up a classifier upon this bag representation for the bag-level prediction, and it is usually referred to as the bag embedding method. Despite the simplicity and being straightforward, instance-level probability pooling is empirically proven to be inferior to the bag embedding counter-part in performance \cite{wang2018revisiting, shao2021transmil}. 

Many of the bag embedding-based models adopt the basic idea of AB-MIL, i.e., the bag embedding (or bag representation) is obtained from the weighting of the features of individual instances, while the latest works of this kind differ in the ways to generate the weighting values, which are typically referred to as attention scores. For example, in the original paper \cite{ilse2018attention}, the attention scores are learned by a side-branch network, in DS-MIL \cite{li2021dual} an attention score is based on the cosine distance between features of an instance and the critical instance, while in Trans-MIL \cite{shao2021transmil}, they are the output of a transformer architecture that encode the mutual-correlations between instances. Essentially, these works are all AB-MIL, and to distinguish, we term the original AB-MIL  \cite{ilse2018attention} as the classic AB-MIL. The main component of our proposed method is also attention-based, but it is not restricted to the way the attention scores are generated. Without loss of generality, we adopt the classic AB-MIL as the base MIL model for each tier in the proposed framework. Please note that altering to other variants of AB-MIL will be straightforward, but is not the main focus of this paper.

\subsection{Grad-based Class Activation Map}

The class activation map (CAM) \cite{zhou2016learning} originally serves as a spatial visualization tool to reveal the locations in an image that correspond to the  classification by a deep learning model. As its generalized version, Grad-CAM (Grad-based Class Activation Map) \cite{selvaraju2017grad} enables the generation of CAM from higher complex architectures of multi-layer perception (MLP). Many works utilized Grad-CAM not only as a powerful tool  for offline model analysis but as an embedded component in the designed deep learning model for various applications. For example, one notable capability of CAM is the target localization of a model trained only with image labels; therefore, it prevails in weakly supervised tasks, such as segmentation \cite{chan2019histosegnet,huang2018weakly, wang2018weakly, kolesnikov2016seed} and detection \cite{wan2018min,wei2018ts2c,zhang2018zigzag}, or even knowledge distillation \cite{wang2021towards}. 

In this paper, we demonstrate that the framework of AB-MIL is a special case of the deep learning architecture for image classification. This finding enables the utilization of the mechanism of Grad-CAM to directly derive the positive probability of an instance under the framework of AB-MIL, and the derivation assists in constructing the proposed framework and also helps with the corresponding analysis.

\section{Method}
\subsection{Revisit Grad-CAM and AB-MIL}

\subsubsection{Grad-CAM} \label{Section_Grad_CAM}

A deep learning model for end-to-end image classification typically comprises of two modules: a deep convolution neural network (DCNN) for high-level feature extraction and a multi-layer perceptron (MLP) for classification. An image is fed into the DCNN to generate the features maps, which become a feature vector by a pooling operation. The feature vector is then forwarded to the MLP for final class probabilities (Fig.\ref{fig_class_gradcam}.(a)). Suppose the final output feature maps from the DCNN is $\bm{U} \in \mathds{R}^{D \times W \times H} $, with $D$ being the number of channel and $W, H$ being the dimensional sizes, respectively. Imposing global average pooling on the feature maps leads to a feature vector that represents the image,

\begin{equation} \label{feat_gap}
\bm{f} =  \underset{W,H}{\text{GAP}}   \left( \bm{U}  \right) \in \mathds{R}^D
\end{equation}

\noindent where $\underset{W,H}{\text{GAP}}\left( \bm{U}  \right)$ denotes the global average pooling with respect to $W,H$, i.e., the $d_{\text{th}}$ element of $\bm{f}$, $f_d = \frac{1}{W H} \sum_{w=1,h=1}^{W,H} U_{w,h}^d  $. Using $\bm{f}$ as the input, the MLP outputs logits $s^c$ for class $c \in \{ 1,2,...,C \}$, whose value indicates the signal strength for the image belonging to class $c$ and the predicted class probability can then be obtained  by soft-max operation accordingly. The class activation map for class $c$ by Grad-CAM is defined as the weighted sum of the feature maps, 

\begin{equation} \label{eq_gradcam}
    \bm{L}^{c} = \sum_{d}^D \beta_d^c \bm{U}^{d}, \ \ \ \ \beta_d^c = \frac{1}{ WH }\sum_{w,h}^{W,H} \left( \frac{ \partial s^c}{\partial U_{w,h}^d} \right), 
\end{equation}

\noindent with $\bm{L}^c \in \mathds{R}^{W \times H}$, and  $L^{c}_{w,h}$ being the magnitude value at location $w,h$ of $\bm{L}^{c}$, indicating how strongly this location tends to be class $c$,

\begin{equation}
   L^{c}_{w,h} = \sum_{d=1}^D \beta_d^c U^{d}_{w,h}
\end{equation}

\subsubsection{Attention-Based Multiple Instance Learning}
Consider a bag of instances $\bm{X} = \{ x_{1}, x_{2}, ... x_{K} \}$ with $K$ being the number of instances in the bag. Each instance $x_{k}, k \in {1,2,...,K}$ holds a latent label  $y_{k}$ ($y_{k}=1$ for positive, or $y_{k}=0$ for negative), which is assumed to be unknown. The goal of MIL is to detect whether there exist at least one positive instance in the bag. The only information revealed for training, however, is the bag label,  which is defined as,

\begin{equation}
Y =
\begin{cases}
1, & \text{if} \ \sum_{k=1}^{K} y_{k} > 0\\
0, & \text{otherwise }
\end{cases}       
\end{equation}

\noindent i.e., the bag is positive if at least one instance in it is positive, or negative otherwise. One straightforward solution for this learning problem is to assign each instance the bag label and accordingly train a classifier, and then apply the max or average pooling operation on the individual instance classifications to obtain the bag-level results \cite{wang2018revisiting}. Another popular strategy is to learn a bag representation $\bm{F}$ from the extracted features of instance in a bag, with which the problem becomes a conventional classification task, i.e., a classifier can be trained upon the bag representations. Empirically, the strategy of bag representation learning is proven to be more efficient than the instance pooling strategy, which we refer to as bag embedding-based MIL. The bag embedding is formulated as,

\begin{equation}
    \bm{F} = \textrm{G} \left( \{ \bm{h}_k \ | \ k=1,2,...,K \} \right),
\end{equation}

\noindent where $\textrm{G}$ is an aggregation function, and $\bm{h}_{k} \in \mathds{R}^{D}$ is the extracted feature for instance $k$. Typically, many  works adopt the attention tactic to obtain the bag representation as follows, 

\begin{equation} \label{eq_weight_feature}
\bm{F} = \sum_{k=1}^{K} a_k \bm{h}_{k} \ \in \mathds{R}^{D},
\end{equation}

\noindent where $a_k$ is the learnable scalar weight for $\bm{h}_{k}$, and $D$ is the dimension of vector $\bm{F}$ and $\bm{h}_k$. The paradigm is shown in  Fig.\ref{fig_class_gradcam}.(b). The attention mechanisms in \cite{ilse2018attention, li2021dual, lu2021data} follow this formulation, therefore they all belong to the category of AB-MIL, but differ in the ways to generate the attention score (weight value) $a_k$. For example, the weight from the classic AB-MIL \cite{ilse2018attention} is defined as,

\begin{equation}
  a_k = \frac{ \exp \{ \bm{w}^{\text{T}} ( \text{tanh}(\bm{V}_1 \bm{h}_k ) \odot \text{sigm}( \bm{V}_2 \bm{h}_k ) )  \}  }{  \sum_{j=1}^{K}  \exp \{ \bm{w}^{\text{T}} ( \text{tanh}(\bm{V}_1 \bm{h}_j ) \odot \text{sigm}( \bm{V}_2 \bm{h}_j ) )  \}   },
\end{equation}

\noindent where $\bm{w}, \bm{V}_1$ and $\bm{V}_2$ are the learnable parameters.

\subsection{Derivation of Instance Probability in AB-MIL}
Despite the better performance of bag embedding-based MIL, it was thought  infeasible to  unravel instance class probability \cite{wang2018revisiting, li2021dual}. In this paper, however, we show that it is possible to derive the predicted probability of each individual instance in a bag under the framework of AB-MIL. The derivation is  rooted in the following proposition,

\begin{prop} \label{prop1}
The paradigm of AB-MIL is a special case of the framework of the classic deep-learning network for image classification.
\end{prop}

\noindent Proof and explanation are in the Supplementary.

Based on Proposition \ref{prop1}, it is safe to apply the mechanism of Grad-CAM to AB-MIL to directly infer the signal strength for an instance to be a certain class. Resembling to Eq.(\ref{eq_gradcam}), the signal strength for instance $k$ to be class $c$ ($c=0$ for negative and $c=1$ for positive) can then be derived as (see Supplementary),

\begin{equation} \label{derive_0}
    L_k^c = \sum_{d=1}^D \beta_d^c \hat{h}_{k,d} , \ \ \beta_d^c = \frac{1}{K} \sum_{i=1}^{K} \frac{ \partial s_c }{ \partial \hat{h}_{k,d} }
\end{equation}

\noindent where $s_c$ is the output logit for class $c$ from the MIL classifier, $ \hat{h}_{k,d} $ is the $d_{\text{th}}$ element of $\hat{\bm{h}}_k$, and $\hat{\bm{h}}_{k} = a_k K \bm{h}_{k} $ with $a_k$ being the attention score for instance $k$ defined in Eq.(\ref{eq_weight_feature}). By applying soft-max, the corresponding probability is then,

\begin{equation} \label{derive_1}
p_k^c = \frac{\exp \left( L_k^c \right) }{ \sum_{t=1}^{C} \exp \left( L_k^t \right) }
\end{equation}

\subsection{Double-Tier Feature Distillation Multiple Instance Learning}

In this section, we present the proposed double-tier feature distillation MIL framework. 

Given $N$ bags (slides), and in each bag there are $K_n$ instances (patches), i.e., $\bm{X}_n= \{x_{n, k} \ | \ k=1,2,...,K_n   \}, n \in \{1,2,...,N\} $, with the ground-truth of a bag being $Y_n$. The corresponding feature of a patch, denoted as $\bm{h}_{n,k}$, is extracted by a backbone network $\textrm{H}$, i.e., $ \bm{h}_{n,k} = \textrm{H}\left(x_{n,k} \right) $. The instances in a slide (bag) are randomly split into $M$ pseudo-bags with approximately even number of instances, $\bm{X}_n = \{ \bm{X}_n^m \ | \   m=1,2,...,M    \}$. A pseudo-bag is assigned the label of its parent bag's label, i.e., $Y_n^m = Y_n$ . In Tier-1, an AB-MIL model, denoted as $\textrm{T}_{1}$, is applied to each pseudo-bag. Suppose the estimated bag probability of a pseudo-bag through the Tier-1 model is $y_n^m$, 

\begin{equation}
	y_n^m = \textrm{T}_1 \big(   \{ \bm{h}_k = \textrm{H} \left(x_k \right) \ | \  x_k \in \bm{X}_n^m \}   \big),
\end{equation}

\noindent The Tier-1 loss function for training using cross entropy is then defined as,

\begin{equation}
   \mathcal{L}_1 =  - \frac{1}{MN} \sum_{n=1,m=1}^{N,M}  Y_n^m \log{y_n^m} + \left(1-Y_n^m \right) \log{ \left( 1-y_n^m \right) }
\end{equation}

\noindent The probabilities of instances in each pseudo-bag can then be derived using Eq.(\ref{derive_0}) and Eq.(\ref{derive_1}). Based on the derived instance probabilities, a feature from each pseudo-bag is distilled, denoted as $\hat{\bm{f}}_{n}^{m}$ for the $m_{\text{th}}$ pseudo-bag of the $n_{\text{th}}$ parent bag. All the distilled features are forwarded into a Tier-2 AB-MIL, denoted as $\textrm{T}_2$, for the inference of the parent bag, 

\begin{equation}
    \hat{y}_n = \textrm{T}_2 \left(    \left\{  \hat{\bm{f}}_{n}^{m}  \ | \ m \in (1,2,...,M) \right\}     \right)
\end{equation}

\noindent The Tier-2 loss function for training $\textrm{T}_2$ is defined as,

\begin{equation}
   \mathcal{L}_2 = -\frac{1}{N} \sum_{n=1}^{N}  Y_n \log \hat{y}_n + (1-Y_n)\log (1-\hat{y}_n),
\end{equation}

\noindent The overall optimization process is then:

\begin{equation}
    \mathcal{L} = \underset{\bm{\theta}_1}{\arg \min} \ \mathcal{L}_1 + \underset{\bm{\theta}_2}{\arg \min} \ \mathcal{L}_2 
\end{equation}

\noindent where $\bm{\theta}_1$ and $\bm{\theta}_2$ are the parameters of $\textrm{T}_1$ and $\textrm{T}_2$, respectively. It should be noted that there is a certain proportion of noise labels for the pseudo bags, as a pseudo bag may not be assigned with at least one positive instance by the random allocation. However, deep neural networks are  resilient to noise labels to some extent. Besides, the noise level can be roughly controlled by the number of pseudo bags in each parent bag, i.e., $M$. We show how $M$'s value will affect the performance of the proposed methods in the ablation study section. 
\\

\noindent Four feature distillation strategies are considered as follows:

\begin{itemize}
\item \textbf{MaxS} Maximum selection: The feature of the instance in a pseudo-bag that achieves the maximum positive probability from the Tier-1 MIL model is selected to forward to the Tier-2 MIL model.

\item \textbf{MaxMinS} MaxMin selection: The features of two instances in a pseudo-bag are distilled and concatenated to forward to the Tier-2 model: the one with the maximum probability and the one with the minimum probability in the pseudo-bag. Such a selection is based on  the consideration that, if only the instance with maximum positive probability in each pseudo-bag are selected (as in the case of MaxS), the decision boundary of the trained Tier-2 model will tend to be pushed forward too tightly to positive samples, and may miss the genuinely positive cases that are similar to the negative ones \cite{xu2019camel}. By introducing the instances of maximum and minimum probabilities at the same time, it gives looser space for the Tier-2 model to generate the parent bag's feature embedding.

\item \textbf{MAS} Maximum attention score selection: The feature of the instance in a pseudo-bag with maximum assigned attention score from the Tier-1 MIL model is distilled to the Tier-2 MIL model.

\item \textbf{AFS} Aggregated feature selection: The feature aggregated from all the instances in a pseudo-bag as in Eq.(\ref{eq_weight_feature}) is forwarded the Tier-2 model. 
\\

\end{itemize}

\noindent We evaluate the performances of all these 4 strategies in the experimental section.

\begin{comment}

\begin{algorithm}
\caption{The procedure of the two-tier feature distillation MIL}\label{alg:cap}
\begin{algorithmic}
	%\Require 
	\State \textbf{Input}: \\
	$N$ bags $\bm{X}_n= \{ x_{n,1}, x_{n,2}, ..., x_{n,I_n}  \}, n \in \{1,2,...,N\} $ 
	\State \textbf{Random Grouping}: Randomly splitting the instances in each bag into $M$ sub-groups: $\bm{X}_n^m = \{  \}$

	%\Ensure $y = x^n$
	\State $y \gets 1$
	\State $X \gets x$
	\State $N \gets n$
	\While{$N \neq 0$}
	\If{$N$ is even}
	\State $X \gets X \times X$
	\State $N \gets \frac{N}{2}$  \Comment{This is a comment}
	\ElsIf{$N$ is odd}
	\State $y \gets y \times X$
	\State $N \gets N - 1$
	\EndIf
	\EndWhile
\end{algorithmic}
\end{algorithm}

\end{comment}

\begin{table*}[ht]
\caption{Results on CAMELYON-16 test set. The subscripts are the corresponding 95\% confidence intervals.  The best ones are  in bold. For DTFD-MIL, the number of pseudo-bags is 5. The flops are measured with the number of instances of a bag being 120, and the instance feature extraction by ResNet-50 is not considered in the presented model sizes and flops.}
\centering
\begin{tabular}{c|cccc|cc}
\hline
\specialrule{.05em}{.05em}{.05em} 
                                        & \multicolumn{3}{c}{$\underline{\textrm{CAMELYON-16}}$}                                                         &  & \multicolumn{2}{l}{} \\
Method                                  & Acc                         & F1                          & AUC                         &  & FLOPs   & Model Size \\ \hline
\multicolumn{1}{l|}{Mean Pooling}       & 0.626$_{(0.616, 0.636)}$ & 0.355$_{(0.346, 0.363)}$ & 0.528$_{(0.518, 0.538)}$ &  & 62.4M   & 524.3K     \\
\multicolumn{1}{l|}{Max Pooling}        & 0.826$_{(0.798, 0.854)}$ & 0.754$_{(0.694, 0.813)}$ & 0.854$_{(0.816, 0.891)}$ &  & 62.4M   & 524.3K     \\
\multicolumn{1}{l|}{RNN-MIL \cite{campanella2019clinical}}            & 0.844$_{(0.818, 0.870)}$ & 0.798$_{(0.791, 0.806)}$ & 0.875$_{(0.873, 0.877)}$ &  & 64.0M   & 1.57M      \\
\multicolumn{1}{l|}{Classic AB-MIL \cite{ilse2018attention}}     & 0.845$_{(0.839, 0.851)}$ & 0.780$_{(0.769, 0.791)}$ & 0.854$_{(0.848, 0.860)}$  &  & 78.1M  & 655.3K     \\
\multicolumn{1}{l|}{DS-MIL  \cite{li2021dual}    }             & 0.856$_{(0.843, 0.869)}$ & 0.815$_{(0.797, 0.832)}$ & 0.899$_{(0.890, 0.908)}$  &  & 117.6M  & 855.7K     \\
\multicolumn{1}{l|}{CLAM-SB \cite{lu2021data} }            & 0.837$_{(0.809, 0.865)}$ & 0.775$_{(0.755, 0.795)}$ & 0.871$_{(0.856, 0.885)}$ &  & 94.8M   & 790.7K     \\
\multicolumn{1}{l|}{CLAM-MB \cite{lu2021data} }            & 0.823$_{(0.795, 0.85)}$  & 0.774$_{(0.752, 0.795)}$ & 0.878$_{(0.861, 0.894)}$ &  & 94.8M   & 791.1K     \\
\multicolumn{1}{l|}{Trans-MIL \cite{shao2021transmil}  }          & 0.858$_{(0.848, 0.868)}$                      &             0.797$_{(0.776, 0.818)}$                &  0.906$_{(0.875, 0.937)}$                       &  &   613.83M      &    2.66M        \\ \hline
\multicolumn{1}{l|}{DTFD-MIL (MaxS)}    & 0.864$_{(0.848, 0.880)}$ & 0.814$_{(0.802, 0.826)}$ & 0.907$_{(0.894, 0.919)}$ &  & 79.4M   & 986.7K     \\
\multicolumn{1}{l|}{DTFD-MIL (MaxMinS)} & 0.899$_{(0.887, 0.912)}$ & 0.865$_{(0.848, 0.882)}$ & 0.941$_{(0.936, 0.944)}$ &  & 80.1M   & 986.7K     \\
\multicolumn{1}{l|}{DTFD-MIL (AFS)}      & \textbf{0.908}$_{(0.892, 0.925)}$ & \textbf{0.882}$_{(0.861, 0.903)}$ & \textbf{0.946}$_{(0.941, 0.951)}$ &  & 79.4M   & 986.7K     \\
\multicolumn{1}{l|}{DTFD-MIL (MAS)}      & 0.897$_{(0.890, 0.904)}$ & 0.864$_{(0.855, 0.873)}$ & 0.945$_{(0.943, 0.947)}$ &  & 79.4M   & 986.7K     \\ \hline
\specialrule{.05em}{.05em}{.05em} 
\end{tabular}
\label{tab_camelyon}
\end{table*}

\begin{table}[t]
	\caption{Results on TCGA lung cancer. The subscripts are the corresponding standard variances. The best ones are  in bold. For DTFD-MIL, the number of pseudo-bags is 8.}
	\begin{tabular}{m{0.17\textwidth}|>{\centering}m{0.07\textwidth}>{\centering}m{0.07\textwidth}>{\centering\arraybackslash}m{0.07\textwidth}}
		\hline
		\specialrule{.05em}{.05em}{.05em} 
		& \multicolumn{3}{c}{$\underline{\textrm{TCGA Lung Cancer}}$}                                      \\ 
		& Acc                  & F1                   & AUC                  \\ \hline
		\footnotesize{Mean Pooling}       & 0.833$_{0.011}$ & 0.809$_{0.012}$ & 0.901$_{0.012}$ \\
		\footnotesize{Max Pooling        } & 0.846$_{0.029}$ & 0.833$_{0.027}$ & 0.901$_{0.033}$ \\
		\footnotesize{RNN-MIL     \cite{campanella2019clinical}         } & 0.845$_{0.024}$ & 0.831$_{0.023}$ & 0.894$_{0.025}$ \\
		\footnotesize{Classic AB-MIL \cite{ilse2018attention}  }  & 0.869$_{0.032}$ & 0.866$_{0.021}$ & 0.941$_{0.028}$ \\
		\footnotesize{DS-MIL   \cite{li2021dual}       }   & 0.888$_{0.013}$ & 0.876$_{0.011}$ & 0.939$_{0.019}$ \\
		\footnotesize{CLAM-SB   \cite{lu2021data}     }    & 0.875$_{0.041}$ & 0.864$_{0.043}$ & 0.944$_{0.023}$ \\
		\footnotesize{CLAM-MB   \cite{lu2021data}    }     & 0.878$_{0.043}$ & 0.874$_{0.028}$ & 0.949$_{0.019}$ \\ 
		\footnotesize{Trans-MIL   \cite{shao2021transmil}    }     & 0.883$_{0.022}$ & 0.876$_{0.021}$ & 0.949$_{0.013}$ \\  \hline
		\footnotesize{DTFD-MIL (MaxS)}    & 0.868$_{0.040}$ & 0.863$_{0.029}$ & 0.919$_{0.037}$ \\
		\footnotesize{DTFD-MIL (MaxMinS)} & \textbf{0.894}$_{0.033}$ & \textbf{0.891}$_{0.027}$ & \textbf{0.961}$_{0.021}$ \\
		\footnotesize{DTFD-MIL (AFS)}      & 0.891$_{0.033}$ & 0.883$_{0.025}$ & 0.951$_{0.022}$ \\
		\footnotesize{DTFD-MIL (MAS)}      & 0.891$_{0.029}$ & 0.890$_{0.021}$ & 0.955$_{0.023}$ \\ \hline
		\specialrule{.05em}{.05em}{.05em} 
	\end{tabular}
	\label{tab_tcga}
\end{table}

\section{Experiments}
In this section, we present the performance of the proposed methods in comparison to other latest MIL works on histopathology WSI, and qualitatively validate the soundness of the derivation of instance probability. We also conduct ablation experiments to further study the proposed methods. More experimental results are presented in the supplementary.

\subsection{Datasets}
We evaluate the proposed methods on two public histopathology WSI datasets: CAMELYON-16 \cite{bejnordi2017diagnostic} and  The Caner Genome Atlas (TCGA) lung cancer. Please refer to Supplementary for the details of these two datasets.

For pre-processing, we apply the OTSU's threshold method to localize the tissue regions in each WSI. Non-overlapping patches of a  size of $256 \times 256$ pixels on the 20X magnification are then extracted from the tissue regions. There are in total 3.7 millions patches from the CAMELYON-16 dataset and 8.3 millions patches from the TCGA Lung Cancer dataset.

\subsection{Implementation Details}
The implementations are described in Supplementary material. For more  details, please refer to the released code.

\begin{comment}
Following \cite{lu2021data, shao2021transmil}, we use the ResNet-50 model \cite{he2016deep} pretrained with ImageNet \cite{deng2009imagenet} as the backbone network to extract an initial feature vector with a dimension of 1024 from each patch, of which the last convolutional module is excluded and a global average pooling is applied to the final feature maps to generate the initial feature vector. The initial feature vector is then reduce to a 512-dimensional feature vector by one fully-connected layer which is served as the ultimate feature representation of a patch. An Adam optimiser \cite{kingma2014adam} with weight decay of 0.0001 is used. All the models are trained for 200 epochs with an initial learning rate of 0.0001, which is reduced to 20\% of itself after 100 epochs. The batch-size is set to be 1, meaning that in each iteration one slide is processed. All experiments were conducted with a NVIDIA V100 GPU.
\end{comment}

\subsection{Evaluation Metrics}

For all the experiments, area under curve (AUC) is the primary performance metric to report, since it is more comprehensive and less sensitive to class imbalance. Besides, the slide-level accuracy (Acc) and F1 score are also considered, which are determined by the threshold of 0.5.

\begin{comment}
\begin{equation}
\begin{aligned}
   \textrm{Acc} & = \frac{ \textrm{TP} + \textrm{TN} }{  \textrm{TP} + \textrm{TN} + \textrm{FP} + \textrm{FN} } \\
   \textrm{F1} & = \frac{ \textrm{TP} }{ \textrm{TP} + \frac{1}{2} \left( \textrm{FP} + \textrm{FN}  \right)  },
\end{aligned}
\end{equation}

\noindent where TP, TN, FP and FN represent true positive, true negative, false positive and false negative, respectively. To determine these four values, the threshold of 0.5 is adopted.
\end{comment}

For CAMELYON-16, the official training set is further randomly split into training and validation sets with a ratio of 9:1. An experiment is running for 5 times, and the mean values of performance metrics on CAMELYON-16 official test set and the corresponding 95\% confidence interval (CI-95) are reported. For TCGA lung cancer, we randomly split the dataset into training, validation, and testing sets with a ratio of 65:10:25 on the patient level. 4-folder cross-validation is adopted, and the mean value of performance metrics of the 4 test folders are reported. Since for each test folder the performances vary significantly, the CI-95 from just 4 values is of less use; therefore, we instead report the corresponding standard variances.

\begin{figure}[h] 
	\includegraphics[width=0.45\textwidth]{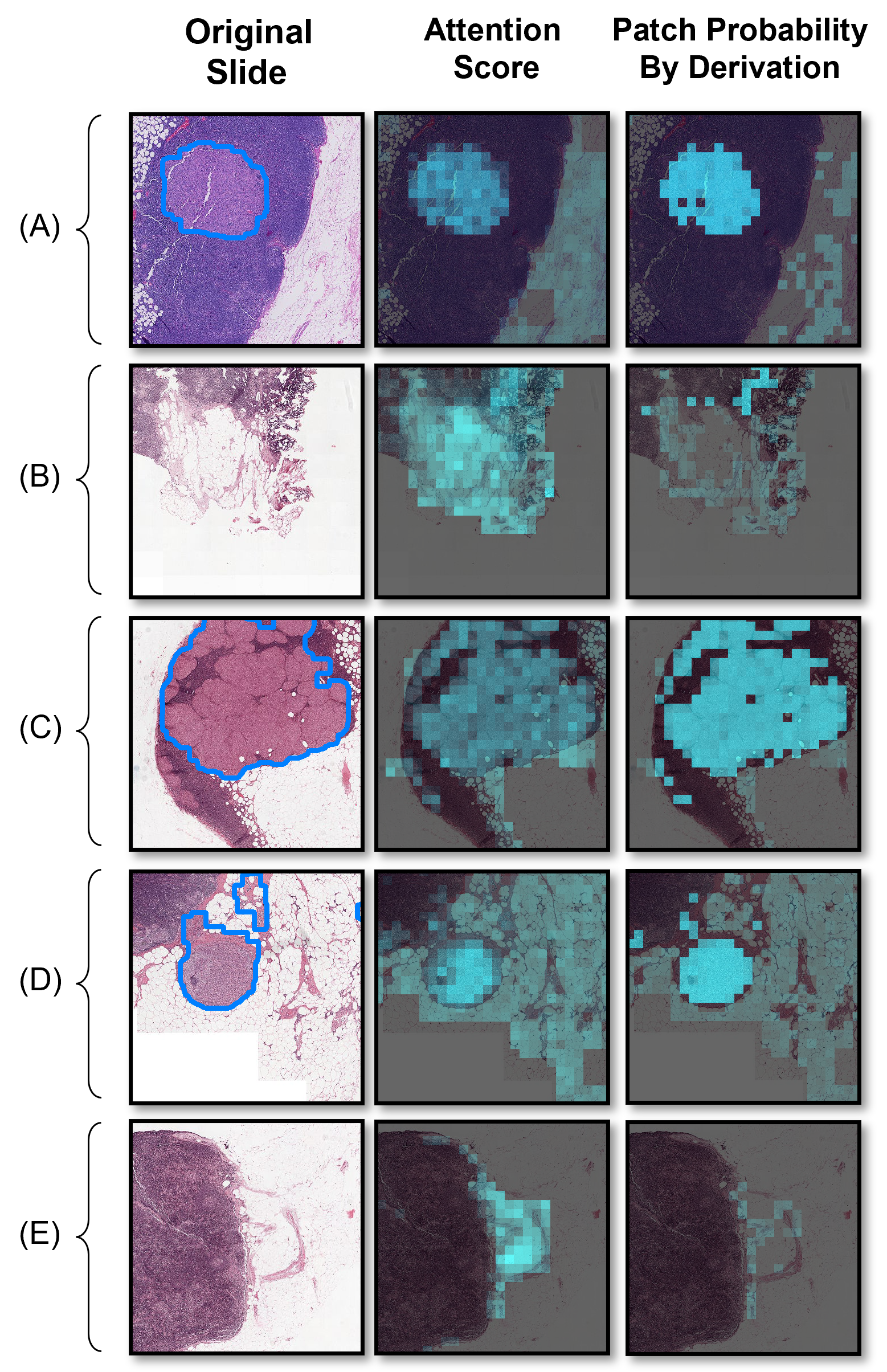}
	\caption{Heat maps of 5 sub-fields of slides by attention score and by the patch probability  derivation, respectively. In the column of `Original Slide', the tumor regions are delineated by the blue lines. In the second and third columns, brighter cyan colors indicate higher probabilities to be tumor for the corresponding locations.  }
	\label{fig_heatmap}
\end{figure}

%The ground-truth slide labels of the 5 cases are positive, negative, positive, positive and negative, respectively.

\subsection{Performance comparison with existing works}
We present the experimental results of the proposed methods on CAMELYON-16 and TCGA lung cancer dataset, in comparison to the following methods: (1) Conventional instance-level MIL, including the Mean-Pooling and Max-Pooling. (2) RNN based RNN-MIL \cite{campanella2019clinical}. (3) The classic AB-MIL \cite{ilse2018attention}. (4) Three variants of AB-MIL, including non-local attention pooling DSMIL \cite{li2021dual}, single-attention-branch CLAM-SB \cite{lu2021data} and multi-attention-branch CLAM-MB \cite{lu2021data}. (5) transformer-based MIL, Trans-MIL \cite{shao2021transmil}.   The results of all the other methods are from the experiments conducted using their official codes under the same settings. As shown in Table.\ref{tab_camelyon}, the proposed models have similar model sizes and computational complexities with the models of other works except for Trans-MIL, while Trans-MIL is significantly larger in model size and computational complexity. 

The results on CAMELYON-16 test set are presented in Tab.\ref{tab_camelyon}, while the results on TCGA lung cancer are shown in Tab.\ref{tab_tcga}. Generally, the instance-level methods (Mean Pooling, Max Pooling) are inferior to the bag embedding-based methods in performances. 

For CAMELYON-16, most slides contain only small portions of tumor over the whole tissue region. Among the proposed DTFD-MIL methods with different feature distillations, the MaxS has the most inferior performances, yet it still outperforms other existing MIL methods except the most recent Trans-MIL. The other 3 DTFD-MIL achieve similar performances, which are significantly better than others. For example, DTFD-MIL(AFS) is at least 4\% better in AUC than  other existing methods.

For TCGA lung cancer, except for DTFD-MIL(MaxS), the proposed methods also achieve leading performances, with DTFD-MIL(MaxMinS) obtaining the best AUC value of 96.1\%. Due to significantly larger tumor regions in positive slides, however, even the instance-level methods perform well on the TCGA lung cancer dataset, resulting in less obvious superiority of the proposed methods over other existing methods. In comparison, for the much more challenging dataset CAMELYON-16, the proposed methods present stronger robustness to the situation of small portions of tumor regions in positive slides.

\begin{comment}
Overall, the idea of pseudo-bags and double-tier MIL indeed provide another plausible perspective to solve the problem of MIL, with superior performances over other latest attention-based methods which exploit the mutual-relations among instances. 
\end{comment}

\subsection{Visualization of Detection Results}

To further explore the proposed instance probability derivation, we train a classic AB-MIL model, and generate the heatmaps of 5 sub-fields of 5 slides from CAMELYON-16. These heatmaps come from (1) normalized attentions scores (attention-based); (2) patch probability derivation (derivation-based) by Eq.(\ref{derive_0}) and Eq.(\ref{derive_1}), respectively. The attention scores directly from the attention module are normalized as $a_k^{'} =  \left( a_k-a_{\textrm{min}} \right) / \left( a_{\textrm{max}} - a_{\textrm{min}} \right)$ \cite{lu2021data, ilse2018attention, li2021dual, shao2021transmil}, where $a_{\textrm{min}}$ and $a_{\textrm{max}}$ are the minimum and maximum attention scores of patches in a slide, respectively. For better presentation, we remove the estimated probabilities of patches in the derivation-based heatmaps (the third row)  whose values are around 0.5 thus contain little information. 

The heatmaps from Fig.\ref{fig_heatmap} demonstrate the better ability of the instance probability derivation to localize the positive activations, compared with the attention scores. Specifically, the positive activations in the heatmaps by instance probability derivation are more consistent and accurate, and present better contrast compared to that of attention scores. Moreover, in the ground-truth negative slides, there are always strong false positive regions in the heatmaps of attention scores, while in the heatmaps from instance probability derivation, most of these regions can be correctly recognized as negative. In the supplementary material, we provide more in-depth analysis about why the instance probability derivation is more efficient for positive activation detection compared to the attention scores.

\subsection{Ablation Study}

\begin{figure}[ht] 
	\centering
	\includegraphics[width=0.43\textwidth]{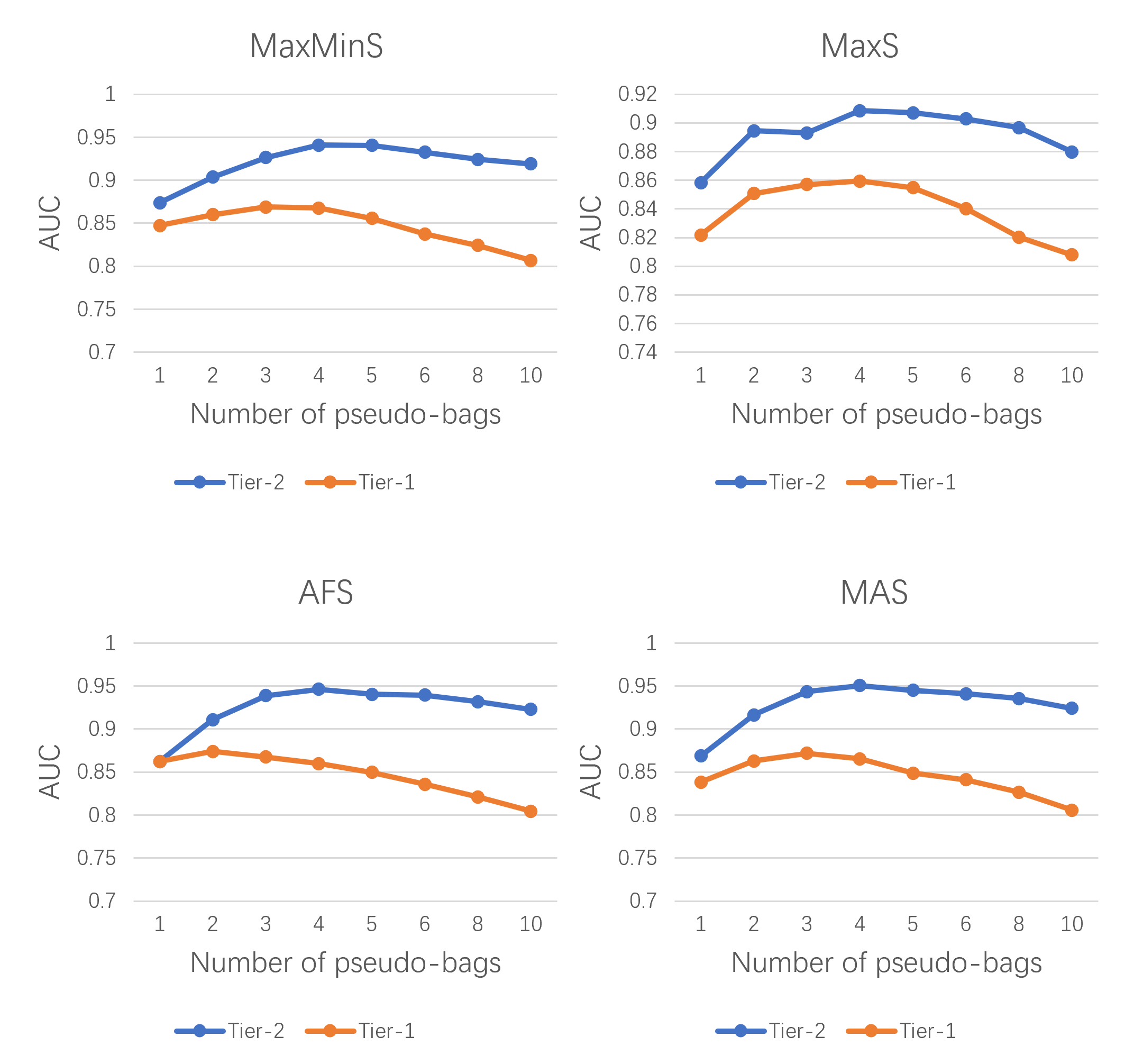}
	\caption{AUC scores of the four feature distillation strategies on CAMELYON-16 test set.}
	\label{fig_ablation_camelyon}
\end{figure}

\begin{figure}[ht]
	\centering
	\includegraphics[width=0.43\textwidth]{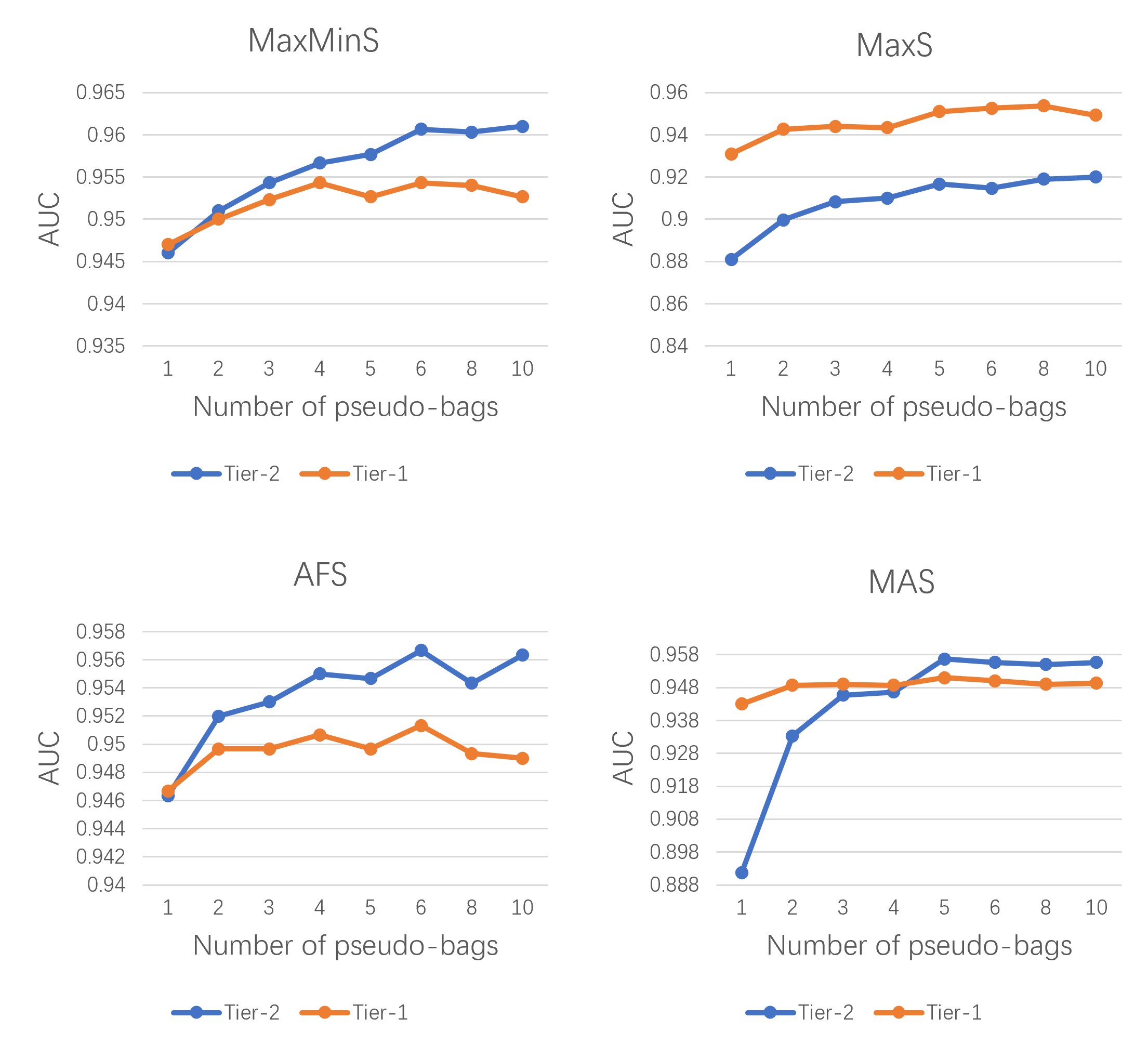}
	\caption{AUC scores of the four feature distillation strategies on TCGA lung cancer dataset.}
	\label{fig_ablation_tcga}
\end{figure}

Fig.\ref{fig_ablation_camelyon} and Fig.\ref{fig_ablation_tcga} present the AUC scores of the proposed methods with respect to different number of pseudo-bags on CAMELYON-16 and TCGA lung cancer datasets, respectively. In each sub-figure, the blue curve represents the Tier-2 MIL model, while the red curve represents the Tier-1 MIL model which directly works on pseudo-bags. 

From these curves, we can summarize:  

\noindent \textbf{(1)}. The pseudo-bag idea is beneficial to both the Tier-1 and Tier-2 MIL models. However, the Tier-1 models are more sensitive to the number of pseudo-bags in CAMELYON-16: the corresponding AUC scores drop dramatically as the number of pseudo-bags increases from 3. In contrast, Tier-1 models are less sensitive to the number of pseudo-bags in TCGA lung cancer dataset, and they even achieve high-level performances with a proper number of pseudo-bags. This phenomenon mainly results from that the tumors are usually minor regions in CAMELYON-16 positive slides while in TCGA lung cancer the situation reverses; therefore, it is highly possible that a pseudo-bag may not be allocated with at least one positive instance from a positive parent bag in CAMELYON-16. This well justifies our initial motivation to build up a second-tier MIL model upon the distilled features from the corresponding pseudo bags, and in general the performances of  Tier-2 models indeed are better than those of  Tier-1 models, especially in CAMELYON-16. 

\noindent \textbf{(2)}. Among the four feature distillation strategies, the DTFD-MIL (MaxS) performance is not comparable to the other three, and on TCGA lung cancer dataset the Tier-2 MIL model is even inferior to the Tier-1 MIL model when MaxS feature distillation is used. It implies that adopting the instances with the highest positive responses to form the representation of the bag is not always the optimal option. This phenomenon is also in accordance with the observation from Fig.\ref{fig_heatmap}, where the strongest activations in a negative slide are from the neutral or even blank regions (corresponding to approximately zero probability of being tumor), instead of the non-tumor tissue regions.

\section{Conclusion}

The first contribution of this paper is the derivation of instance probability under the framework of AB-MIL, and we qualitatively demonstrate the derived instance probability is a more reliable metric over the widely-used attention scores for the positive region detection. We then propose the DTFD-MIL, which utilizes the idea of pseudo bags and double-tier MIL. The derivation of instance probability serves for the feature distillation in DTFD-MIL. The experimental results demonstrate that the proposed DTFD-MIL indeed  provides a new perspective to solve the MIL problem with superior performances, rather than utilizing mutual-instance relations as in other latest works. Finally, we also expect that the derivation of instance probability will be served as a useful tool for developing related MIL models or for the related analysis in future works, just as the role it plays in the proposed DTFD-MIL in this paper.
\\

\noindent \textbf{\large{Acknowledgement}} H. Zhang and Y. Meng thank the China Science IntelliCloud Technology Co., Ltd for the studentships. The TCGA Lung cancer dataset is from the TCGA Research Network: https://www.cancer.gov/tcga.

\clearpage

%%%%%%%%% REFERENCES
{\small
\bibliographystyle{ieee_fullname}
\bibliography{egbib}
}

\end{document}